\documentclass[letterpaper]{article} 
\usepackage{aaai25}  
\usepackage{times}  
\usepackage{helvet}  
\usepackage{courier}  
\usepackage[hyphens]{url}  
\usepackage{graphicx} 
\usepackage{natbib}  
\usepackage{caption} 
\usepackage{algorithm}
\usepackage{algorithmic}

\usepackage{amssymb}
\usepackage{amsmath}
\usepackage{multirow}
\usepackage{subcaption}
\newcommand{\sys}{FedVCK}

\usepackage{newfloat}
\usepackage{listings}
\DeclareCaptionStyle{ruled}{labelfont=normalfont,labelsep=colon,strut=off} 
\floatstyle{ruled}
\newfloat{listing}{tb}{lst}{}
\floatname{listing}{Listing}
\title{FedVCK: Non-IID Robust and Communication-Efficient Federated Learning via Valuable Condensed Knowledge for Medical Image Analysis}
\author{
    Guochen Yan\textsuperscript{\rm 1,3,4},
    Luyuan Xie\textsuperscript{\rm 2,3,4}, 
    Xinyi Gao\textsuperscript{\rm 5}, 
    Wentao Zhang\textsuperscript{\rm 6},\\
    Qingni Shen\textsuperscript{\rm 2,3,4}\thanks{Corresponding author: qingnishen@ss.pku.edu.cn}, 
    Yuejian Fang\textsuperscript{\rm 2,3,4}, 
    Zhonghai Wu\textsuperscript{\rm 2,3,4}\thanks{Corresponding author: wuzh@pku.edu.cn}
}
\affiliations{
    \textsuperscript{\rm 1}School of Computer Science, Peking University, Beijing, China\\
    \textsuperscript{\rm 2}School of Software and Microelectronics, Peking University, Beijing, China\\
    \textsuperscript{\rm 3}PKU-OCTA Laboratory for Blockchain and Privacy Computing, Peking University, Beijing, China\\
    \textsuperscript{\rm 4}National Engineering Research Center for Software Engineering, Peking University, Beijing, China\\
    \textsuperscript{\rm 5}The University of Queensland, Brisbane, Australia \\
    \textsuperscript{\rm 6}Center for Machine Learning Research, Peking University, Beijing, China\\

     Guochen Yan@outlook.com, 2201110745@stu.pku.edu.cn, xinyi.gao@uq.edu.au, wentao.zhang@pku.edu.cn, qingnishen@ss.pku.edu.cn, fangyj@ss.pku.edu.cn, wuzh@pku.edu.cn
}

\usepackage{bibentry}

\begin{document}

\maketitle

\begin{abstract}
Federated learning has become a promising solution for collaboration among medical institutions. However, data owned by each institution would be highly heterogeneous and the distribution is always non-independent and identical distribution (non-IID), resulting in client drift and unsatisfactory performance. Despite existing federated learning methods attempting to solve the non-IID problems, they still show marginal advantages but rely on frequent communication which would incur high costs and privacy concerns. In this paper, we propose a novel federated learning method: \textbf{Fed}erated learning via \textbf{V}aluable \textbf{C}ondensed \textbf{K}nowledge (\sys). We enhance the quality of condensed knowledge and select the most necessary knowledge guided by models, to tackle the non-IID problem within limited communication budgets effectively. Specifically, on the client side, we condense the knowledge of each client into a small dataset and further enhance the condensation procedure with latent distribution constraints, facilitating the effective capture of high-quality knowledge. During each round, we specifically target and condense knowledge that has not been assimilated by the current model, thereby preventing unnecessary repetition of homogeneous knowledge and minimizing the frequency of communications required. On the server side, we propose relational supervised contrastive learning to provide more supervision signals to aid the global model updating. Comprehensive experiments across various medical tasks show that \sys\ can outperform state-of-the-art methods, demonstrating that it's non-IID robust and communication-efficient. 
\end{abstract}

%

\section{Introduction}

Federated learning has become increasingly attractive since it allows collaborative training among sensitive institutions without direct data sharing. However, in reality, each medical institution would have its specialization, and the private data are highly related to the regional demographic characteristics. The data owned by each client are non-independent and identical distribution (non-IID), exhibiting significant data heterogeneity and imbalance. Under this scenario, federated learning methods suffer a global model with unsatisfactory performance due to the model divergence and client drift phenomenon~\cite{li2019convergence,li2022federated}. Meanwhile, frequent communication between heterogeneous and dispersed institutions would incur high communication costs, delays, and complex administrative procedures, with increasing privacy and safety risks~\cite{zhu2019deep,mothukuri2021survey}. A non-IID robust and communication-efficient federated learning method is desired.

\begin{table}[t]
\centering
\setlength{\tabcolsep}{1mm}
\small
\begin{tabular}{c|c|c}
\hline
Method & \textbf{Syn. Data Quality} & \textbf{Knowledge Selection} \\ \hline
FedGen & calibrate classifiers & heuristic   \\
FedMix & distorted sample & repeated  \\
FedGAN & match single sample & repeated   \\ 
DFRD & infidelity & heuristic   \\ 
FedDM & only match final feature & repeated \\ 
DESA & only match final feature & repeated   \\ \hline
\sys & \textbf{latent dist. constraints} & \textbf{model-guided selection} \\
\hline
\end{tabular}
\caption{Representative data-centric methods' problems of 1) synthesis data quality and 2) knowledge selection in the synthesis. `heuristic' indicates they adopt heuristic diversity loss with no relation to the need of models. `repeated' indicates they do nothing and thus select data with repeated knowledge in synthesis. In contrast, we adopt latent distribution constraints and model-guided selection respectively.}
\label{tab:summary_chanllenge}
\end{table}

\begin{figure*}[t]
\centering
\begin{subfigure}{0.49\textwidth}
    \centering
    \includegraphics[width=1.0\columnwidth]{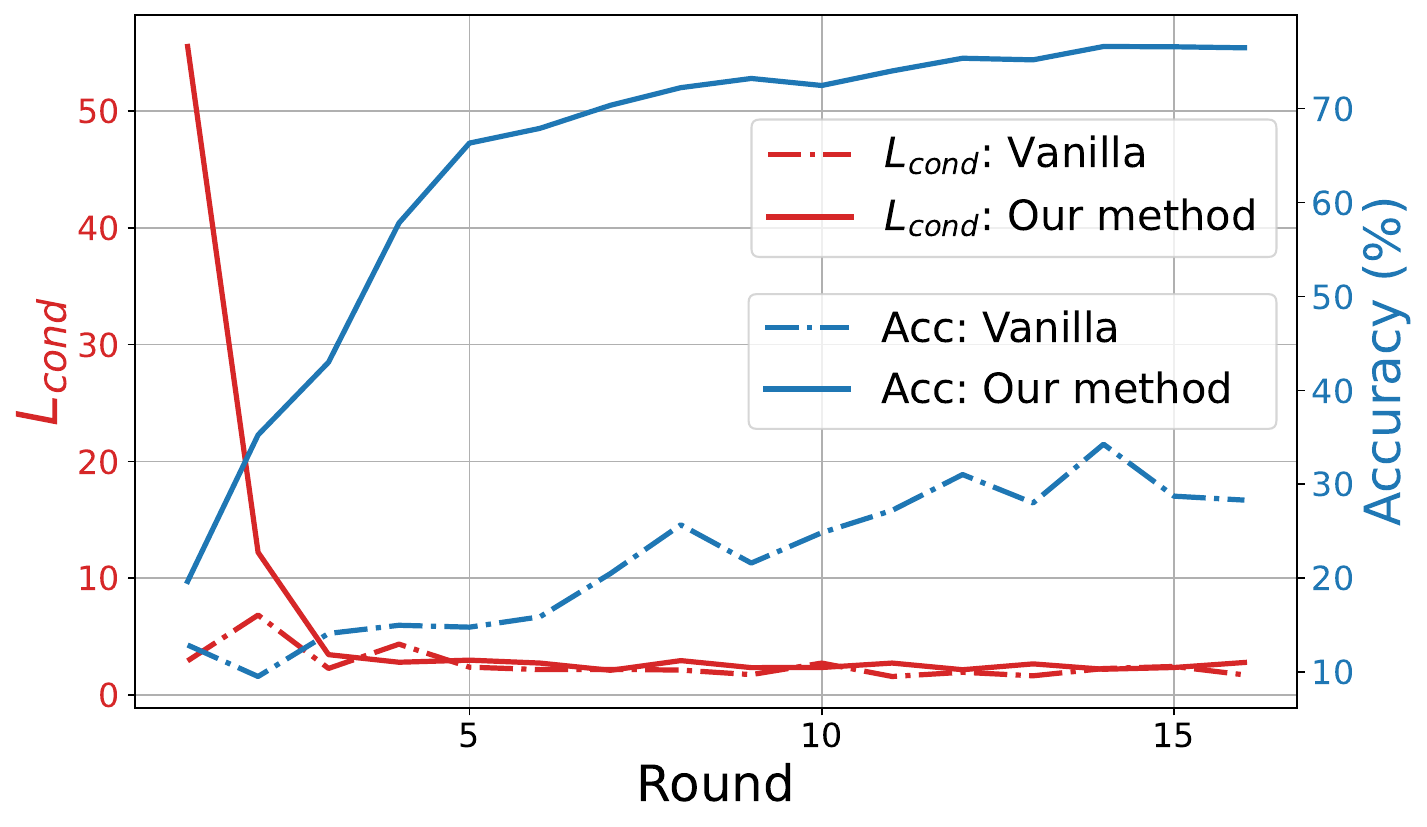}
    \caption{We use the distribution matching-based method to condense knowledge on the OrganC dataset. Without our latent distribution constraints (Vanilla, dashed line), $L_{cond}$ would be easily reduced in each round but the model's performance struggles to improve with the condensed knowledge, demonstrating the low-quality problem of vanilla methods.}
    \label{fig:organc_demo}
\end{subfigure}
\hfill
\begin{subfigure}{0.49\textwidth}
    \centering
    \includegraphics[width=0.98\columnwidth]{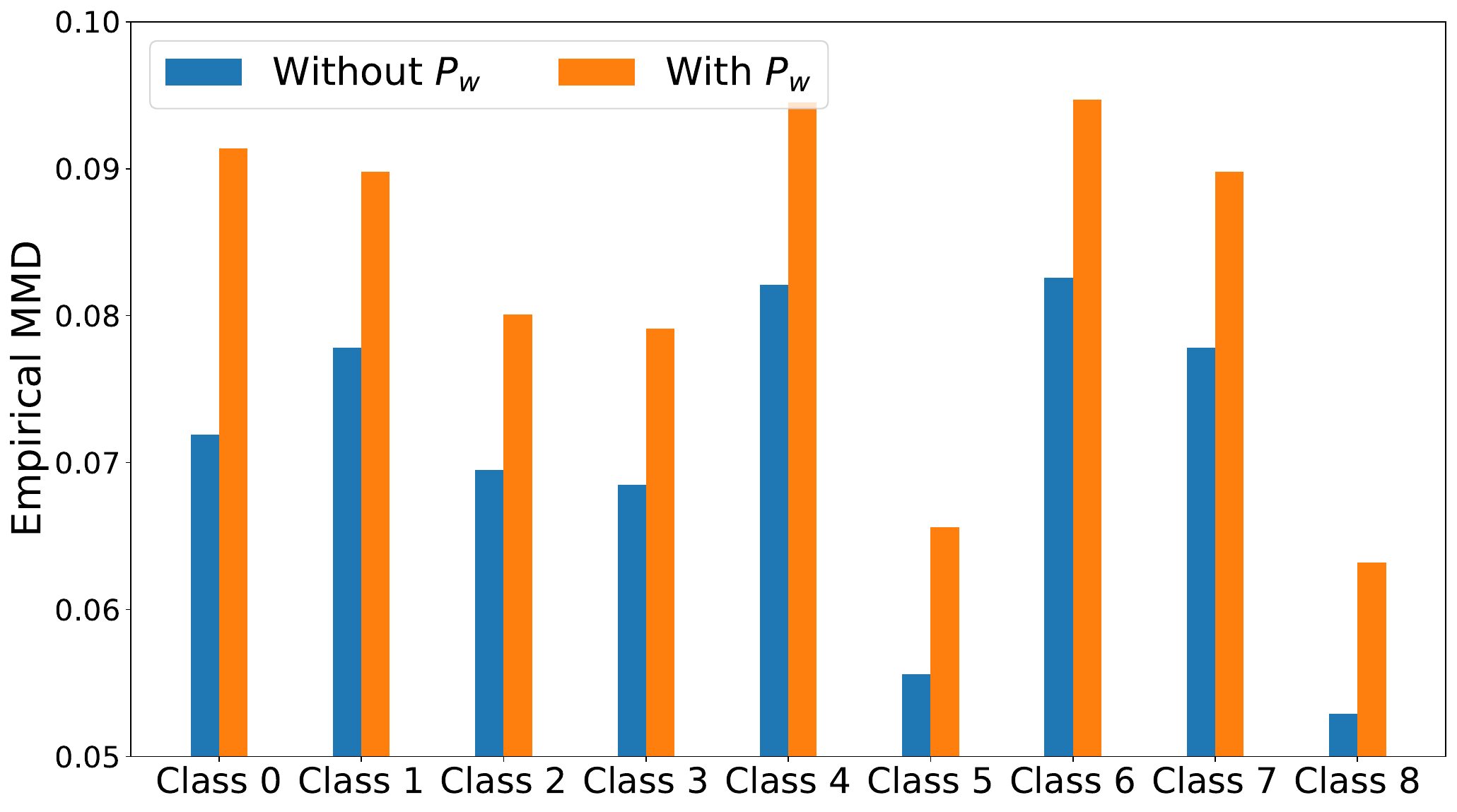}
    \caption{We measure the average MMD of condensed knowledge class-wisely between adjacent rounds on the Path dataset. Greater MMD indicates a larger distribution difference. The vanilla selection causes more knowledge repetition between rounds. Our model-guided selection ($P_w$) ensures that the condensed knowledge between adjacent rounds exhibits greater differences.}
    \label{fig:path_mmd_gap}
\end{subfigure}
\caption{Illustration of the low synthesized data quality problem in Figure~(a) and repeated knowledge problem in Figure~(b).}
\end{figure*}

Many federated learning methods are proposed to cope with non-IID problems by modifying local training process~\cite{li2020federated,li2021model,zhou2023fedfa} or global aggregation process~\cite{chen2020fedbe,lin2020ensemble,zheng2023federated}. However, they are mainly \textbf{model-centric}. They focus on mitigating model parameter-level divergence indirectly under a typical paradigm of local training and global aggregation, leading to marginal advantages in performance and communication costs under severe non-IID scenarios~\cite{li2022federated}.

The model divergence originates from the data divergence~\cite{zhao2018federated}, thus mitigating the data-level divergence would be more essential to tackle the non-IID problems. Recently, various \textbf{data-centric} federated learning methods attempted to share virtual or synthesized data to mitigate data divergence. They synthesize diverse objectives including latent features~\cite{zhu2021data}, approximated real data~\cite{li2022federatedgan,zhu2022federated,yoon2021fedmix}, inverted data~\cite{zhang2022fine,wang2024dfrd}, condensed data~\cite{xiong2023feddm,huang2024overcoming,wang2024aggregation} and so on. However, under severe non-IID scenarios, these methods still face problems because of: 1) \textbf{low synthesized data quality}. For instance, mix-up would distort data~\cite{verma2019manifold}. The inverted data is of infidelity with biased models. And the advanced dataset condensation cannot effectively extract subtle and meaningful knowledge which we demonstrate in Figure~\ref{fig:organc_demo}. These low-quality data would fail to guide the model training; 2) \textbf{repeated knowledge}. The data are randomly selected to synthesize virtual data, and their value and importance to the current model are not considered. Thus, the knowledge contained tends to be homogeneous and unnecessarily repeated (see Figure~\ref{fig:path_mmd_gap}), thus cannot effectively update the model after several rounds. We summarize problems in the above two aspects of representative data-centric federated learning methods in Table~\ref{tab:summary_chanllenge}. Additionally, some synthesis methods would incur privacy concerns, and most methods are \textbf{not communication-efficient}. They still face challenges to achieve a satisfactory performance under limited communication rounds in non-IID scenarios.

Motivated by the above limitations, we propose a novel data-centric \textbf{Fed}erated learning method via \textbf{V}aluable \textbf{C}ondensed \textbf{K}nowledge (\sys). Our method includes two parts, valuable knowledge condensation on the client side and relational supervised learning-aided updating on the server side. Specifically, we condense each client's knowledge into a small dataset. To ensure condensing high-quality knowledge, we propose latent distribution constraints to better capture subtle and meaningful knowledge in latent spaces. To minimize redundancy in each round of knowledge condensation, we explicitly measure the missing knowledge of the current model and select the most necessary knowledge in condensation on each client. On the server side, we identify the hard negative classes for each class and propose a relational supervised contrastive learning to enhance the supervision signals during model updating. Due to the balanced, high-quality, unrepeated, and necessary condensed knowledge, the training of the global model is insulated from the effects of non-IID problems and can achieve enhanced performance within limited communication rounds (e.g. 10). Moreover, our method only condenses task-related high-level knowledge with random noise initialization, thereby facilitating privacy protection. Our main contributions are summarized as follows:

\begin{itemize}
    \item We propose a novel data-centric federated learning method: \sys, for collaborative medical image analysis. \sys\ is robust to severe non-IID scenarios and communication efficient with valuable knowledge.
    \item On the client side, we propose model-guided selection to sample the most needed knowledge each round to avoid unnecessary repetition. We also propose latent distribution constraints to enhance the quality of knowledge.
    \item On the server side, we identify the hard negative classes and propose relational supervised contrastive learning to enhance supervised learning in model updating.
    \item We conduct comprehensive experiments and results show that our method achieves better predictive performance, especially under limited communication budgets. We also conduct experiments to verify the privacy-preserving ability and generality of our method.
\end{itemize}

\section{Related Works}
The data owned by each client is typically highly heterogeneous and does not follow an independent and identical distribution. Under severe non-IID scenarios, models trained on clients tend to be highly biased and divergent, a phenomenon known as client drift. Aggregating these biased and divergent client models at the server often results in suboptimal performance. Many model-centric methods focus on modifying local training process, such as introducing regularization or contrastive terms to reduce divergence~\cite{li2020federated,acar2021federated,li2021model,Xie_MHpFLGB_MICCAI2024,xie2024mhpflid, Xie_pFLFE_MICCAI2024} or improving aggregation process~\cite{lin2020ensemble,chen2020fedbe,zheng2023federated}. They try to alleviate the client drift from the model parameter level.

The model divergence originates from the data divergence~\cite{zhao2018federated}. Directly reducing the difference in data distribution would reduce the model divergence fundamentally. Recently, data-centric federated learning methods have drawn attention since they can synthesize and then share virtual synthesized data to mitigate the non-IID problem in a data-centric manner. Besides that FedGen~\cite{zhu2021data} which generates virtual representation, various format data are synthesized on the server or clients. FedMix~\cite{yoon2021fedmix} broadcasts the mix-up data to approximate the real data. FedGAN~\cite{nguyen2021federated} and SDA-FL~\cite{li2022federatedgan} train and share GANs to imitate real data to support COVID-19 detection. FedFTG~\cite{zhang2022fine} and advanced version DFRD~\cite{wang2024dfrd} use model inversion~\cite{yin2020dreaming} to generate data for knowledge distillation. FedDM~\cite{xiong2023feddm} condenses the knowledge to update the global model. DESA~\cite{huang2024overcoming} distills anchor data and broadcasts them to enable mutual regularization and distillation among clients.

\begin{figure*}[ht]
\centering
\includegraphics[width=0.98\textwidth]{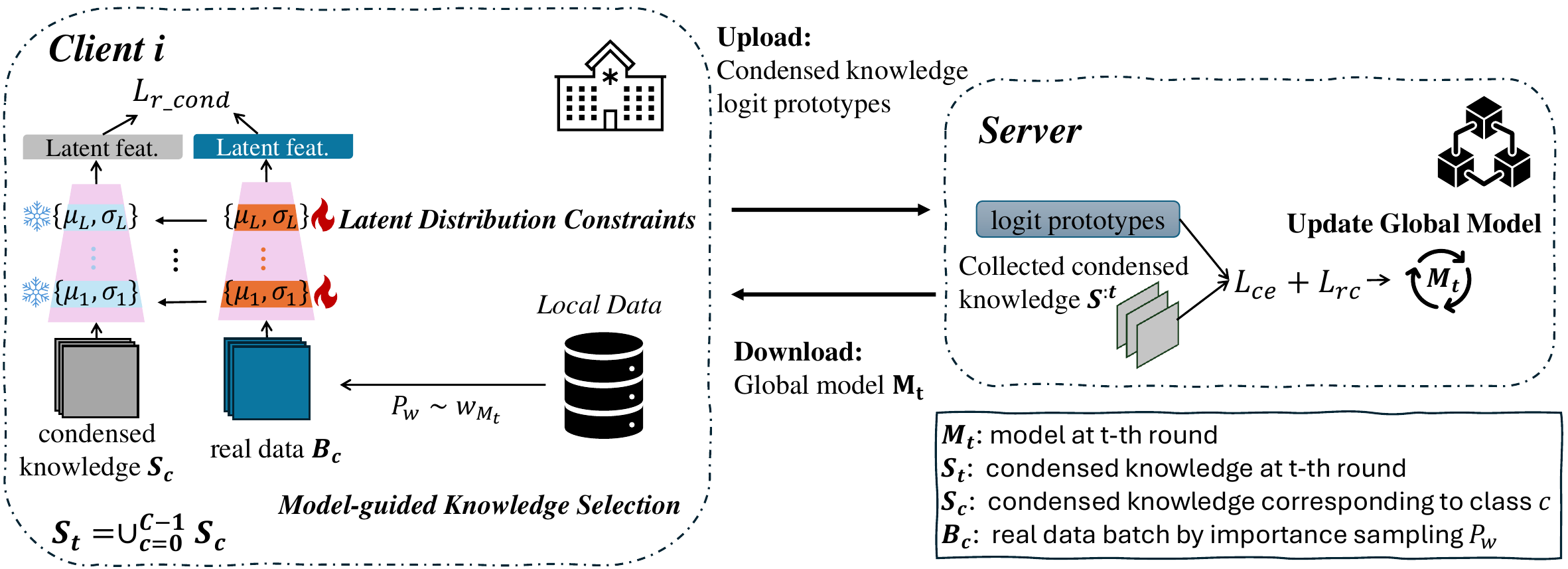}
\caption{Overview of \sys. \textbf{On the client side}, we sample local data by importance sampling guided by the current model and then impose latent distribution constraints in optimization. We upload the condensed knowledge dataset and logit prototypes to the server. \textbf{On the server side}, we use cross entropy loss and relational contrastive loss to update the global model.}
\label{fig:overview}
\end{figure*}

\section{Proposed Method}

\subsection{Overview}
The overview of \sys\ is shown in Figure~\ref{fig:overview}. In short, it consists of two parts: valuable knowledge condensation on the client side and relational supervised learning-aided updating on the server side. On the client side, we borrow distribution matching techniques in dataset condensation and optimize the learnable dataset to condense knowledge from local data. To ensure quality, we record dynamic distribution statistics of the local data batch in each encoder layer and replace the statistics during embedding learnable knowledge as fixed constraints, which could force the latent distribution of the condensed knowledge to capture subtle and meaningful knowledge of different levels. To minimize redundancy in each round of condensation, we explicitly measure the prediction error on each sample and select the data on which the model performs poorly. We consider such data critical as it contains knowledge not yet captured by the current model. By focusing more on these important samples, the condensation process ensures that the condensed knowledge complements the global model’s missing capabilities. On the server side, we collect the condensed knowledge dataset and train the global model with supervised learning and relational contrastive learning. We first identify hard negative classes for each class where the global model tends to mispredict by uploaded logit prototypes. Then we use supervised contrastive learning in a bootstrap manner to draw the features of the same class closer to their prototypes and push the features away from their hard negative classes' prototypes. We will introduce our designs in detail in the following sections.




\subsection{Preliminary: Dataset Condensation}
The objective of dataset condensation~\cite{wang2018dataset, yu2023dataset, gao2024graph1, gao2024graphsurvey} is to condense knowledge from a large dataset into a small learnable dataset, which could be used to train models to achieve comparable performance. Distribution matching is an advanced method widely used in dataset condensation.

\paragraph{Distribution matching.} The intuition behind is to optimize a small dataset $\mathcal{S}$ to match the latent feature distribution of local data $\mathcal{T}$ by minimizing the distance to the latent features of local data with maximum mean discrepancy (MMD)~\cite{Gretton_Borgwardt_Rasch_Schölkopf_Smola_2012,Zhao_Bilen_2023}:
\begin{equation}
\mathop{\arg\min}\limits_{\mathcal{S}} \sup_{\|\psi_{\theta}\|_{\mathcal{H}}\leq 1}(\mathbb{E}[\psi_{\theta}(\mathcal T)] - \mathbb{E}[\psi_{\theta}(\mathcal S)]),
\label{eq:theory_mmd}
\end{equation}
where $\mathcal{H}$ is reproducing kernel Hilbert space (RKHS), $\psi_{\theta}$ is the shared embedding function to map the input to its latent feature, parameterized by a multi-layer encoder. In practice, We minimize the estimated empirical MMD loss by class-wisely align the latent feature distributions to optimize $\mathcal{S}$:
\begin{equation}
    L_{cond} = \sum_{c=0}^{C-1}\|\frac{1}{|B_c|}\sum_{x_i \in B_c}\psi_{\theta}(x_i) - \frac{1}{|\mathcal{S}_c|}\sum_{s_i \in \mathcal{S}_c}\psi_{\theta}(s_i)\|^2,
\label{eq:linear mmd loss}
\end{equation}
where $C$ is the number of classes, $\mathcal{T}_c$ is the local data with class $c$, $B_{c}$ is a batch randomly sampled from $\mathcal{T}_c$ with a uniform distribution, and $\mathcal{S}_c$ is the knowledge dataset corresponding to class $c$. To enable the high-order estimation, we choose to align the latent feature distributions in an RKHS with kernel $\mathcal K$~\cite{DBLP:conf/aaai/0003LW0G24}, and minimize the following empirical MMD as condensation loss:
\begin{equation}
    L_{cond} = \sum_{c=0}^{C-1}\sum_{B_{c} \overset{P_u}{\sim} \mathcal T_c} \hat{\mathcal{K}}_{B_{c},B_{ c}}+\hat{\mathcal{K}}_{\mathcal{S}_c,\mathcal{S}_c}-2\hat{\mathcal{K}}_{B_c, \mathcal{S}_c}, \\
\label{eq:mmd loss}
\end{equation}
where $\hat{\mathcal{K}}_{X,Y} =\frac{1}{|X|\cdot|Y|}\sum_{i=1}^{|X|}\sum_{j=1}^{|Y|}\mathcal{K}(\psi_{\theta}(x_i), \psi_{\theta}(y_j))$, $\{x_i\}_{i=1}^{|X|}\sim X, \quad \{y_j\}_{j=1}^{|Y|}\sim Y$.
The kernel function $\mathcal K$ can be a linear kernel, inner-product kernel, or Gaussian kernel.

\paragraph{Knowledge initialization.} There are several manners to initialize learnable knowledge dataset $\mathcal{S}$ whose format is the same as real data. To best prevent the privacy leak of the local data, we choose random Gaussian noise $\mathcal{N}(0,1)$ to initialize the knowledge dataset, making the knowledge can only be condensed by matching the latent distributions. The condensed knowledge dataset $\mathcal{S}$ would contain no individual and privacy information in pixel space and the adversary can hardly infer the membership from the condensed knowledge datasets~\cite{dong2022privacy}.

\subsection{Latent Distribution Constraints}
By optimizing $\mathcal{S}$ with $L_{cond}$ in Eq.~\ref{eq:linear mmd loss} or Eq.~\ref{eq:mmd loss}, the condensed knowledge dataset $\mathcal{S}$ can replace the real data to effectively train models. However, it's challenging to ensure the condensation quality when condensing the knowledge from local data into a small random noise-initialized dataset, because random noises have no prior about the local data. Moreover, the condensation loss ($L_{cond}$) is not sufficient to guide the learning of subtle meaningful knowledge since the latent features of $\mathcal{S}$ can take shortcuts to over-fit the latent features of local data. To assess the deficiency of the vanilla optimization process, we show in Figure~\ref{fig:organc_demo} that while $L_{cond}$ is easily reduced at each round, it can not effectively condense meaningful knowledge to improve performance effectively. Additionally, matching the distribution of the final representation of the encoder neglects the previous intermediate latent feature distributions. 

To enhance the effectiveness and ensure consistent distribution of all latent features, we transfer dynamic distribution statistics (mean and variation) from the local real data to the condensed knowledge data during the condensation procedure. Compared to~\cite{yin2020dreaming}, our approach does not require the addition of an extra loss term, enabling a more flexible and efficient condensation procedure. Specifically, we first record the distribution statistics of each layer $\{\{\mu_1, \sigma_1\}, ..., \{\mu_L, \sigma_L\}\}$ with a $L$ layer encoder when embedding a batch of local data. Then the statistics during embedding condensed data are replaced and fixed with recorded statistics:
\begin{equation}
\begin{aligned}
    s_i^{(1)} &= Norm(\psi_{1}(s_i), \{\mu_1,\sigma_1\}), \\
    s_i^{(2)} &= Norm(\psi_{2}(s_i^{(1)}), \{\mu_2,\sigma_2\}), \\
    &...\\
    \psi_{\theta}(s_i) = s_i^{(L)} &= Norm(\psi_{L}(s_i^{(L-1)}, \{\mu_L,\sigma_L\}),
\end{aligned}
\end{equation}
where $\psi_{l}$ is $l$-th layer of encoder $\psi_{\theta}$ and $Norm$ denotes batch normalization~\cite{ioffe2015batch}. The distribution statistics constraint could force the optimization process to consider intermediate latent distributions and prevent it from taking shortcuts. Thus the quality of condensed knowledge can be largely improved.

\subsection{Model-guided Knowledge Selection}
If we uniformly sample real data from local data to conduct condensation, the condensed knowledge in each round will be repeated and homogeneous, dominated by simple and easy-to-learn knowledge. It would be less beneficial to further improve the performance of the global model. However, from the model perspective, we find that the importance of the knowledge contained in each sample varies. The current global model may perform well on some local data but lacks the ability to make correct predictions on others, exposing that the current model lacks some knowledge. The data containing the missing knowledge would be more important at this round and it's better to focus on condensing knowledge from these data to complement the model knowledge. Specifically, we first measure the importance of each data sample explicitly by model prediction error as $t$-th round:
\begin{equation}
    w_{\mathbf{M}_t}(x_i) = \frac{1}{1+e^{-err_t(x_i) + b}},
\label{eq:importance}
\end{equation}
where the $err_t(x_i)$ refers to the prediction error on $x_i$ of current model $
\mathbf{M}_t$ at $t$-th round and $b$ is a constant to control the scale range. The higher the prediction error, the more important $x_i$ would be. Here we adopt the cross-entropy loss as the prediction error with the current model $\mathbf{M}_t$:
\begin{equation}
\begin{aligned}
    err_t(x_i) = L_{ce}(y_i, \mathbf{M}_t(x_i)).
\end{aligned}
    \label{eq:vanilla_pred_err}
\end{equation}

The current model is usually not well-trained and may over-fit on limited uploaded condensed knowledge, the distribution of loss would be skewed and less calibrated to reflect the proper importance relation. We propose the self-ensemble of the current model $\mathbf{M}_t$ and previous model $\mathbf{M}_{t-1}$ to smooth and regularize the distribution of loss. Thus the desired knowledge can be condensed progressively. we refine the prediction error of Eq.~\ref{eq:vanilla_pred_err} as:
\begin{equation}
\begin{aligned}
    \tilde{err}_t(x_i) &= L_{ce}(y_i, \alpha \mathbf{M}_t(x_i) + (1-\alpha) \mathbf{M}_{t-1}(x_i)), \\
\end{aligned}
\end{equation}
where $\alpha$ is a hyper-parameter. With refined prediction error, we can calculate the refined importance in Eq.~\ref{eq:importance} and replace the uniform sampling $P_u$ in Eq.~\ref{eq:mmd loss} with importance sampling $P_w$ conditioned on model $\mathbf{M}_{t}$ and $\mathbf{M}_{t-1}$:
\begin{equation}
\begin{aligned}
    P_w(x_i | \mathbf{M}_t, \mathbf{M}_{t-1}) = \frac{w_{\mathbf{M}_t}(x_i)}{\sum_{x_j} w_{\mathbf{M}_t}(x_j)}.
\end{aligned}
\end{equation}

We then refine the condensation loss of Eq.~\ref{eq:mmd loss} as :
\begin{equation}
\begin{aligned}
    L_{r\_cond} = \sum_{c=0}^{C-1} \hat{\mathcal{K}}_{B_{c}^{P_w},B_{ c}^{P_w}}+\hat{\mathcal{K}}_{\mathcal{S}_c,\mathcal{S}_c}-2\hat{\mathcal{K}}_{B_{c}^{P_w}, \mathcal{S}_c}, 
\end{aligned}
\end{equation}
where data in each batch $B_{c}^{P_w}$ is sampled based on $P_w$. Thus the condensation process in the $t$-th round can be regarded as a biased variant of Eq.~\ref{eq:theory_mmd} where the where the expectation over $\mathcal{T}$ is replaced by a weighted expectation under $P_w$:
\begin{equation}
\mathop{\arg\min}\limits_{\mathcal{S}} \sup_{\|\psi_{\theta}\|_{\mathcal{H}}\leq 1}(\mathbb{E}_{P_w}[\psi_{\theta}(\mathcal T)] - \mathbb{E}[\psi_{\theta}(\mathcal S)]),
\end{equation}

Note that the current model $\mathbf{M}_t$ is dynamically updating. We measure the importance and derive the importance sampling $P_w$ at each round. Thus, the condensed knowledge in each round can continue to transition from known knowledge to missing knowledge. The global model could complement its ability at each round, making its performance improve consistently.

\subsection{Relational Prototype-wise Contrastive Learning}
On the server side, we calculate the global logit prototype for each class and identify their hard negative classes. Afterward, prototype-wise contrastive learning is deployed to facilitate the discrimination between classes.

At the $t$-th round, besides the condensed knowledge dataset, each client $k$ uploads the logit prototypes $\{\mathbf{p_0}^{k, t}, ..., \mathbf{p_{C-1}}^{k, t}\}$ calculated by the global model as:
\begin{equation}
\mathbf{p_c}^{k, t} = \frac{1}{N_{c, k}}\sum_{i}^{N_{c,k}}f_{M_{t-1}}(x_{i, c, k}),
\end{equation}
where the $x_{i, c, k}$ denotes the local data with class $c$ in client $k$, and $f_{M_{t-1}}$ denotes the current model without the last softmax layer at the beginning of the $t$-th round. Then we aggregate these prototypes uploaded from each client into global logit prototypes $\{\mathbf{p_0}^t, ..., \mathbf{p_{C-1}}^t\}$:
\begin{equation}
\mathbf{p_c}^t = \frac{1}{|\mathcal{T}_c|}\sum_{k}^{N}|\mathcal{T}_{c, k}|\mathbf{p_c}^{k, t},
\end{equation}
where $N$ denotes the number of clients, $|\mathcal{T}_c|$ denotes the total number of data of class $c$, and $|\mathcal{T}_{c, k}|$ denotes the number of data of class $c$ in client $k$. With global logit prototypes, we can derive the Top-K hard negative classes for class $c$ as :
\begin{equation}
HN(c) = \{j_1, j_2, ...j_K\} = \mathop{\arg topK}_{j\neq c}\; \mathbf{p_c}[j],
\end{equation}
where $HN(c)$ contains the class indices with Top-K values in prototype vector $\mathbf{p_c}$ except $c$. We recognize $HN(c)$ as hard negative classes' indices set for class $c$ since the global always predicts a higher probability on these classes and tends to mis-classify. To amplify discrimination ability of the global model, it would be more effective to push features of class $c$ away from that of $HN(c)$. Note that the global logit prototypes may change across rounds with the updated global model, $HN(c)$ would also change adaptively.

We also calculate feature prototypes $\{\mathbf{f_0}^t, ..., \mathbf{f_{C-1}}^t\}$ with condensed knowledge datasets on the server at $t$-th round:
\begin{equation}
\mathbf{f_c}^t = \frac{1}{|\mathcal{S}_c^{:t-1}|}\sum_{s_i \in \mathcal{S}_c^{0, ..., t-1}}\psi_{\theta}^{t-1}(s_i),
\end{equation}
where $\mathcal{S}_c^{: t-1}$ is the accumulated knowledge dataset of class $c$ uploaded before $t$-th round, and $\psi_{\theta}^{t-1}$ is the encoder of the global model $\mathbf{M}_{t-1}$ at the very beginning of $t$-th round.

With hard negative classes set $HN(c)$ and feature prototypes, inspired by SimSiam~\cite{chen2021exploring}, we propose relational supervised contrastive learning with prototypes in a bootstrap manner:
\begin{small}
\begin{equation}
\begin{aligned}
    L_{rc} = \sum_{(s_i, c_i) \in S^{:t}} -\log \frac{\exp{(h(\psi_{\theta}^{t}(s_i)) \cdot \mathbf{f_{c_i}}^t / \tau)}}{\sum_{c_j \in HN(c_i)} \exp{(h(\psi_{\theta}^{t}(s_i)) \cdot \mathbf{f_{c_j}}^t / \tau)}},
\end{aligned}
\end{equation}
\end{small}

where $h$ is a learnable projector similar with ~\cite{chen2021exploring, grill2020bootstrap} and $\tau$ is the temperature hyper-parameter. Then we update the global model along with cross-entropy loss at $t$-th round with:
\begin{equation}
    L_{update} = L_{ce}(\mathcal{S}^{:t}) + L_{rc},
\end{equation}
where $L_{ce}$ denotes the cross-entropy loss with condensed knowledge datasets and the relational supervised contrastive learning offers more supervision signals in model updating.

\section{Experiments}

\paragraph{Datasets.} We evaluate the performance of our proposed \sys\ on 4 medical tasks, which contain 5 datasets with different modalities from~\cite{medmnistv1, medmnistv2}: 1) Colon Pathology, we adopt the Path dataset, 2) Retinal OCT scans, we adopt the OCT dataset, 3) Abdominal CT scans, we adopt the OrganS and OrganC dataset, 4) Chest X-Ray, we adopt the Pneumonia dataset. To validate the generality, we also select CIFAR10~\cite{krizhevsky2009learning}, STL10~\cite{coates2011analysis}, and ImageNette~\cite{imagenette} datasets. Our selected datasets enjoy a wide range of modalities and resolutions from 28$\times$28 to 224$\times$224 and detailed introductions about datasets are shown in the Appendix.

\paragraph{Baselines.} We compare \sys\ with nine federated learning methods including both model-centric methods (FedAvg, FedProx, and MOON) and data-centric methods (FedGen, FedMix, FedGAN, DFRD, FedDM, and DESA). We summarize rationale of the baseline selection and their synthesis objectives and methods in Table~1 in Appendix.

\begin{table*}[t]
\centering
\small
\begin{tabular}{c|ccc|cc|ccc|cc}
\hline
$\beta$ & \multicolumn{5}{c|}{0.05} &  \multicolumn{5}{c}{0.02} \\ \hline
Model & \multicolumn{3}{c|}{ConvNet} & \multicolumn{2}{c|}{ResNet18} &  \multicolumn{3}{c|}{ConvNet} & \multicolumn{2}{c}{ResNet18} \\ \hline
Acc(\%) & \textbf{Path} & \textbf{OCT} & \textbf{OrganS} & \textbf{OrganC} & \textbf{Pneumonia} & \textbf{Path} & \textbf{OCT} & \textbf{OrganS} & \textbf{OrganC} & \textbf{Pneumonia} \\
\hline
FedAvg & 46.34 & 62.00 & 66.27 & 70.38 & 69.87 & 43.72 & 26.54 & 60.70 & 65.87 & 63.14 \\
FedProx & 61.34 & 62.50 & 69.14 & 71.80 & 69.07 & 40.15 & 32.20 & 67.76 & 60.94 & 62.50 \\
MOON & 51.91 & 56.10 & 52.33 & 71.82 & 62.50 & 50.88 & 29.90 & 62.84 & 61.81 & 62.50 \\ \hline
FedGen & 42.03 & 53.25 & 59.90 & 49.65 & 60.37 & 39.87 & 34.74 & 47.06 & 37.63 & 58.43 \\
FedGAN & 54.40 & 56.80 & 71.76 & OOM & 75.32 & 54.37 & 25.20 & 70.34 & OOM & 62.50 \\
FedMix & 35.78 & 48.90 & 62.10 & 60.63 & 62.50 & 31.50 & 29.30 & 54.65 & 29.22 & 62.50 \\
DFRD & 37.44 & 31.50 & 39.80 & OOM & OOM & 14.01 & 34.20 & 37.93 & OOM & OOM \\
FedDM & 73.97 & 61.70 & 71.37 & 35.60 & 75.80 & 73.64 & 62.20 & 69.46 & 18.20 & 68.75 \\
DESA & 33.37 & 47.00 & 69.98 & 54.16 & 61.38 & 66.41 & 35.20 & 67.32 & 39.46 & 62.50 \\ \hline
\textbf{\sys} & \textbf{80.36} & \textbf{68.30} & \textbf{73.23} & \textbf{79.52} & \textbf{86.70} & \textbf{81.10} & \textbf{68.20} & \textbf{72.90} & \textbf{79.04} & \textbf{84.62} \\
\hline
\end{tabular}
\caption{Overall predictive accuracy comparison on medical datasets. We test our method and baselines under two non-IID scenarios: $Dir(0.05)$ and $Dir(0.02)$. For datasets with 224$\times$224 image sizes, we adopt the ResNet18 model. \textbf{Bold} numbers indicate the best accuracy results. `OOM' indicates out-of-memory.}
\label{tab:adequate_medical}
\end{table*}

\begin{table*}[ht]
\centering
\small
\begin{tabular}{c|ccc|cc|ccc|cc}
\hline
$\beta$ & \multicolumn{5}{c|}{0.05} &  \multicolumn{5}{c}{0.02} \\ \hline
Model & \multicolumn{3}{c|}{ConvNet} & \multicolumn{2}{c|}{ResNet18} &  \multicolumn{3}{c|}{ConvNet} & \multicolumn{2}{c}{ResNet18} \\ \hline
Acc(\%) & \textbf{Path} & \textbf{OCT} & \textbf{OrganS} & \textbf{OrganC} & \textbf{Pneumonia} & \textbf{Path} & \textbf{OCT} & \textbf{OrganS} & \textbf{OrganC} & \textbf{Pneumonia} \\
\hline
FedAvg & 18.55 & 53.70 & 36.92 & 29.31 & 62.50 & 12.84 & 25.00 & 39.32 & 25.13 & 35.74 \\
FedProx & 56.94 & 49.70 & 48.76 & 50.60 & 63.78 & 40.15 & 28.70 & 49.60 & 46.13 & 62.50 \\
MOON & 49.40 & 32.00 & 45.69 & 33.43 & 62.50 & 26.96 & 25.00 & 42.60 & 25.83 & 62.50 \\ \hline
FedGen & 34.47 & 27.60 & 45.75 & 24.89 & 58.00 & 37.20 & 25.00 & 37.79 & 18.77 & 57.50 \\
FedGAN & 19.89 & 48.00 & 55.65 & OOM & 64.74 & 32.32 & 25.00 & 44.67 & OOM & 62.50 \\
FedMix & 28.97 & 31.60 & 57.35 & 32.46 & 62.50 & 28.48 & 25.00 & 42.37 & 17.73 & 62.50 \\
DFRD & 22.40 & 25.00 & 39.05 & OOM & OOM & 10.45 & 25.00 & 29.93 & OOM & OOM \\
FedDM & 72.76 & 61.70 & 71.32 & 31.66 & 73.27 & 70.39 & 62.20 & 69.33 & 15.12 & 62.50 \\
DESA & 29.48 & 36.90 & 66.00 & 50.72 & 38.78 & 43.62 & 27.80 & 66.42 & 39.46 & 37.50 \\ \hline
\textbf{\sys}  & \textbf{78.52} & \textbf{66.41} & \textbf{72.65} & \textbf{71.39} & \textbf{83.01} & \textbf{78.61} & \textbf{65.90} & \textbf{71.68} & \textbf{71.89} & \textbf{82.48} \\
\hline
\end{tabular}
\caption{Predictive accuracy comparison on medical datasets under limited communication budgets.}
\label{tab:limited_medical}
\end{table*}

\paragraph{Configuration.} Following the commonly used setting, we simulate non-IID scenarios with Dirichlet distribution $Dir(\beta)$ among 10 clients where $\beta$ is 0.05 and 0.02 to simulate severe non-IID scenarios. We adopt the ConvNet~\cite{gidaris2018dynamic} and ResNet18~\cite{he2016deep}. We set the size of the condensed knowledge dataset $\mathcal{S}$ to $p$\% of the original dataset size, where $p$ is selected from \{1, 2, 5\} according to different datasets. We initialize $S$ from $\mathcal{N}(0,1)$. Hyper-parameters in each method are tuned as suggested in the original papers. We run all experiments with NVIDIA Geforce RTX 3090 GPU and report the mean results among three runs. More details are introduced in the Appendix.

\subsection{Performance Under Non-IID Scenarios}

\paragraph{Overall performance.} We evaluate all methods' overall performance under non-IID scenarios in Table~\ref{tab:adequate_medical}, assuming the communication budgets are adequate (100 communication rounds). We can observe that model-centric federated learning methods struggle with mediocre performance. Some data-centric methods (e.g. FedGen and DFRD) perform worse. We find this is because the poor synthesis quality and poor global model hinder each other and cause a vicious circle. On OrganC and Pneumonia datasets, FedGAN and DRFD face the out-of-memory problem. Clients in FedGAN must train and upload huge generators and discriminators and the server in DFRD must maintain ensemble models and huge generators. Most data-centric baselines degrade hardly on the two datasets since capturing subtle and meaningful knowledge in larger sizes is harder. Our method successfully condenses knowledge with high quality and high necessity for the global model, thus showing advantages over all datasets' baselines.

\paragraph{Under limited communication budgets.} Since communication budgets are usually limited in reality, a method that can achieve high performance within a few communication rounds is more desired. We compare the performance of our method and baselines within 10 communication rounds under non-IID scenarios. The experimental results are shown in Table~\ref{tab:limited_medical}. We can observe that all baselines cannot achieve satisfactory performance within limited rounds, while our method consistently outperforms others on all datasets. The performance is relatively close to the overall performance in Table~\ref{tab:adequate_medical} and would not be significantly affected by more severe non-IID ($\beta=0.02$), demonstrating that our method is not only communication-efficient but also robust to non-IID.


\subsection{Performance Analysis}

\begin{table}[htb]
\centering
\begin{tabular}{c|cc|cc}
\hline
Acc(\%) & \textbf{Path} & \textbf{OrganS} & \textbf{Path} & \textbf{OrganS}  \\ 
\hline
w.o. all & 72.76 & 71.32 & 73.97 & 71.37  \\
w.o. $L_{rc}$ + $P_w$ & 73.04 & 71.74 & 76.48 & 72.13    \\ 
w.o. $L_{rc}$ & 74.52 & 71.93  & 78.56 & 72.40    \\ \hline
\textbf{\sys} & 78.52 & 72.65 & 80.36 & 73.23  \\
\hline
\end{tabular}
\caption{Ablation study on medical datasets. The left part of the table is the performance under limited communication rounds and the right part is the overall performance.}
\label{tab:ablation_study}
\end{table}

\paragraph{Ablation study.}
We conduct ablation study to evaluate the effectiveness of our designs. The experimental results are shown in Table~\ref{tab:ablation_study}. Besides, we measure the empirical MMD of the condensed knowledge between two adjacent rounds with or without model-guided selection in Figure~\ref{fig:path_mmd_gap}. We can note that with model-guided selection, the condensed knowledge between adjacent rounds exhibits greater MMD values, reflecting that it can avoid repeated knowledge and force the optimization process to condense more heterogeneous and model-specific knowledge. We also study the impact of the size of learnable knowledge dataset in the Appendix. Larger size would have more capacity but increase optimization difficulty and communication overhead.

\paragraph{Communication analysis.}
Our method is communication efficient from two aspects. From the perspective of communication rounds, we have demonstrated our method can quickly achieve satisfactory performance under limited budgets in Table~\ref{tab:limited_medical}. From the perspective of upload communication costs, we quantify the actual per-round upload communication costs of all clients in Table~\ref{tab:commnucation_msg_quatity}. Our method's per-round uploading costs are less than that of model-centric federated learning. More analysis about the communication and full experimental results are shown in the Appendix.


\begin{table}[tb]
\centering
\setlength{\tabcolsep}{1mm}
\small
\begin{tabular}{cc|ccc}
\hline
\multicolumn{2}{c|}{\textbf{Method}}   & \textbf{Path} & \textbf{OrganS} & \textbf{Pneumonia} \\ \hline
\multicolumn{2}{c|}{FedMix, DRFD} & \multirow{4}{*}{12.13 MB} & \multirow{4}{*}{12.20 MB} & \multirow{4}{*}{426.15 MB} \\ 
\multicolumn{2}{c|}{FedAvg, FedProx} &   &  &  \\ 
\multicolumn{2}{c|}{MOON, FedGen} & &  &  \\ 
\multicolumn{2}{c|}{DESA} & &  &  \\ \hline
\multicolumn{2}{c|}{FedGAN} & 178.85 MB & 178.69 MB & 2349.40 MB \\ \hline
\multicolumn{2}{c|}{\sys, FedDM} & \textbf{2.04 MB} & \textbf{0.52 MB} & \textbf{11.77 MB}  \\ 
\multicolumn{2}{c|}{$p$\%} & 1\% & 5\% & 5\% \\ \hline
\end{tabular}
\caption{Per-round upload communication costs. $p$\% indicates the size of the condensed knowledge dataset as a proportion of the size of the original dataset.}
\label{tab:commnucation_msg_quatity}
\end{table}

\subsection{Privacy Analysis}
To practically test whether the condensed knowledge would leak individual privacy, we conduct the membership inference attack following an advanced method: LiRA~\cite{carlini2022membership} and compare \sys\ with the model-centric federated learning method (e.g. FedAvg). We attack uploaded models or condensed knowledge from clients and record the AUC of ROC on each client with a balanced test set. Since the MIA task is a binary classification, we set the minimum AUC to 0.5 and mark it as a total defense if the AUC of a client is less than or equal to 0.5. We calculate the max and mean AUC of all clients and defense rate (the proportion of clients achieving total defense) in Table~\ref{tab:MIA results}. The results show that our method better preserves privacy than FedAvg, and enables more total defense cases. In addition, since our method needs fewer communication rounds, privacy can be further protected from potential temporal-based MIA~\cite{zhu2024evaluating}.

\begin{table}[tb]
\centering
\setlength{\tabcolsep}{1mm}
\small
\begin{tabular}{c|c|c|c}
\hline
 \textbf{Method} & \textbf{Max AUC $\downarrow$} & \textbf{Mean AUC $\downarrow$} & \textbf{Defense Rate $\uparrow$}  \\ \hline
\textbf{FedAvg} & 0.556 & 0.529 & 10\%  \\
\textbf{\sys} & \textbf{0.544} & \textbf{0.514} & \textbf{50\%} \\ \hline

\end{tabular}
\caption{The AUC results of MIA experiment on OrganS dataset. $\downarrow$ means the lower, the better. $\uparrow$ means the opposite.}
\label{tab:MIA results}
\end{table}

\begin{table}[tb]
\centering
\setlength{\tabcolsep}{1mm}
\begin{tabular}{c|cc|cc|cc}
\hline
Acc(\%) & \multicolumn{2}{c|}{\textbf{CIFAR10}} & \multicolumn{2}{c|}{\textbf{STL10}} & \multicolumn{2}{c}{\textbf{ImageNette}} \\ \hline
$\beta$ & 0.05 & 0.02 & 0.05 & 0.02 & 0.05 & 0.02  \\
\hline
FedAvg & 52.13 & 52.01 & 46.03 & 39.60 & 47.21  & 38.17  \\
FedProx & 57.60 & 53.39  & 42.93 & 42.88 & 52.66 & 32.84 \\
MOON & 46.63 & 42.03 & 38.08 & 38.42 & 35.26 & 21.32 \\ \hline
FedGen & 39.36 & 32.71 & 38.11 & 37.44 & 51.92 & 41.89 \\
FedGAN & 55.79 & 53.86 & 51.84 & 50.03 & 50.80 & 40.31 \\
FedMix & 42.28 & 43.97 & 46.56 & 42.88 & 50.80 & 39.41 \\
DFRD & 52.07 & 37.53 & 31.60 & 21.03 & 33.20 & 16.20 \\
FedDM & 54.75 & 50.47 & 54.90 & 51.62 & 52.25 & 43.82 \\
DESA & 53.90 & 48.19 & 46.74 & 37.33 & 42.29 & 28.90 \\ \hline
\textbf{\sys}  & \textbf{62.96} & \textbf{60.56} & \textbf{57.04} & \textbf{56.89} & \textbf{62.76} & \textbf{61.73}\\
\hline
\end{tabular}
\caption{Overall predictive accuracy comparison on natural datasets. We adopt the ConvNet model by default.}
\label{tab:adequate_natural}
\end{table}

\subsection{Extend to Natural Datasets} 
To validate the generality of our method, we also extend our evaluation on natural datasets. The natural datasets contain various colored objects with more significant inter-class differences. The overall experimental results are shown in Table~\ref{tab:adequate_natural}. Our method still outperforms others consistently, which demonstrate a boarder generality of our method. Full experiments about the predictive performance, communication cost, and privacy-preserving are listed in the Appendix.

\section{Conclusion and Discussion}
In this paper, we propose a novel data-centric federated learning method, \sys, for collaborative medical image analysis. \sys\ can tackle the non-IID problem in a communication-efficient manner. Specifically, \sys\ adaptively selects the most necessary knowledge with the guidance of current models, and condenses it into a small knowledge dataset with latent distribution constraints to enhance the quality. The condensed knowledge can effectively update the global model with the help of relational supervised contrastive learning. Our method generally outperforms state-of-the-art methods in non-IID scenarios, especially under limited communication budgets. Further work is to extend to more data modalities such as 3D CT and to adopt advanced techniques to improve the effectiveness and efficiency of condensation.

\appendix

\section{Acknowledgments}
This work is supported by the National Key R\&D Program of China under Grant No.2022YFB2703301.

\bigskip

\bibliography{aaai25}

\clearpage

\setcounter{secnumdepth}{0} 

%




\section{Datasets and Configurations}
We evaluate methods on two categories of datasets: medical datasets and natural datasets. For medical datasets, we adopt 5 widely used datasets with different modalities and image sizes. 
\begin{itemize}
    \item Path consists of 107,180 histopathologic images of colorectal cancer, collected from hematoxylin and eosin-stained histological slides. The dataset is divided into 89,996 training images, 10,004 validation images, and 7,180 test images. The images are 28×28 pixels and contain nine types of tissues: tumor epithelium, simple stroma, complex stroma, immune cells, debris, mucus, smooth muscle, normal colon mucosa, and cancer-associated stroma.
    \item OCT is based on optical coherence tomography images of the retina. This dataset includes 109,309 images categorized into four classes: choroidal neovascularization, diabetic macular edema, drusen, and normal. The dataset is split into 97,477 training images, 10,832 validation images, and 1,000 test images. The images are gray-scale and resized to 28$\times$28 pixels.
    \item OrganS comprises 25,221 abdominal CT images, labeled into 11 classes representing different organ segments. The dataset is split into 13,940 training images, 2,452 validation images, and 8,829 test images. The images are 28$\times$28 pixels in size and are used for multi-class classification tasks.
    \item OrganC contains 23,660 abdominal CT images, categorized into 11 classes. The dataset is divided into 13,000 training images, 2,392 validation images, and 8,268 test images. We use the large-size version. All samples are 224$\times$224 pixels in size. It is utilized for multi-class classification tasks.
    \item Pneumonia is derived from a dataset of 5,856 chest X-ray images, with labels for pneumonia and normal cases. The dataset includes 4,708 training images, 524 validation images, and 624 test images. We use the large-size version. All samples are 224$\times$224 pixels in size.
\end{itemize}

For natural datasets, we adopt 3 widely used datasets and follow the standard splits:
\begin{itemize}
    \item CIFAR10 consists of 60,000, 32$\times$32 color images in 10 classes, with 6,000 images per class. There are 50,000 training images and 10,000 test images. The classes are airplane, automobile, bird, cat, deer, dog, frog, horse, ship, and truck.
    \item STL10 is a 10-class color image dataset. Each class contains 1,300 images. We resize each sample to 32$\times$32, similar to CIFAR10. The classes include airplane, bird, car, cat, deer, dog, horse, monkey, ship, and truck.
    \item ImageNette is a subset of the larger ImageNet dataset, containing 10 classes and around 10,000 images. The classes include tench, English springer, cassette player, chain saw, church, French horn, garbage truck, gas pump, golf ball, and parachute. We resize each sample to 64$\times$64.
\end{itemize}
We adopt the standard preprocessing and splits on all datasets and report the mean accuracy among three runs. We adopt the ConvNet~\cite{gidaris2018dynamic} model and adopt ResNet18~\cite{he2016deep} model for datasets with 224$\times$224 image sizes. We set the size of the condensed knowledge dataset $\mathcal{S}$ to $p$\% of that of the original dataset, where we select $p$ from \{1, 2, 5\} according to different datasets. We initialize $S$ from $\mathcal{N}(0,1)$. We set the local epochs as 10. We tune the learning rate of model training and knowledge learning from 0.001 to 0.01. Other hyper-parameters in each method are tuned as suggested in the original papers. We implement all methods with Pytorch 2.1 on Ubuntu 20.04 equipped with an NVIDIA Geforce RTX 3090 GPU and Intel(R) Xeon(R) CPU E5-2680 v4.

\begin{table}[ht]
\centering
\setlength{\tabcolsep}{1mm}
\begin{tabular}{c|c|c}
\hline
Method & \textbf{Synthesis Objective} & \textbf{Synthesis Method}  \\ \hline
FedGen & latent feature  &  generator at server  \\
FedMix & original sample & mix up   \\
FedGAN & original sample  & GAN at clients \\ 
DFRD & original sample  & model inversion  \\ 
FedDM & latent knowledge  & distribution matching   \\ 
DESA & anchor data  & distribution matching  \\ 
\hline
\end{tabular}
\caption{Summary of synthesis objectives and methods of representative data-centric federated learning methods.}
\label{tab:summary_baseline}
\end{table}


\section{Summary of Data-centric Baselines}
Various data-centric federated learning methods synthesize various objectives with different methods. The synthesis objectives including 1) latent features~\cite{zhu2021data}, synthesized by a generator on the server, 2) original data~\cite{nguyen2021federated, li2022federatedgan, zhu2022federated}, synthesized by generative models such as GANs, 3) data mixture~\cite{yoon2021fedmix}, synthesized by simply mix up the samples and labels of a few data, 4) model inverted data~\cite{zhang2022fine, wang2024dfrd}, synthesized by model inversion, 5) latent knowledge~\cite{xiong2023feddm, wang2024aggregation, huang2024overcoming} by distribution matching. Considering different synthesis objectives and methods, We select FedGen, FedMix, FedGAN, DFRD, FedDM and DESA as representative baselines. We summarize them in Table~\ref{tab:summary_baseline}.

\begin{figure}[htbp]
\centering
\includegraphics[width=0.9\columnwidth]{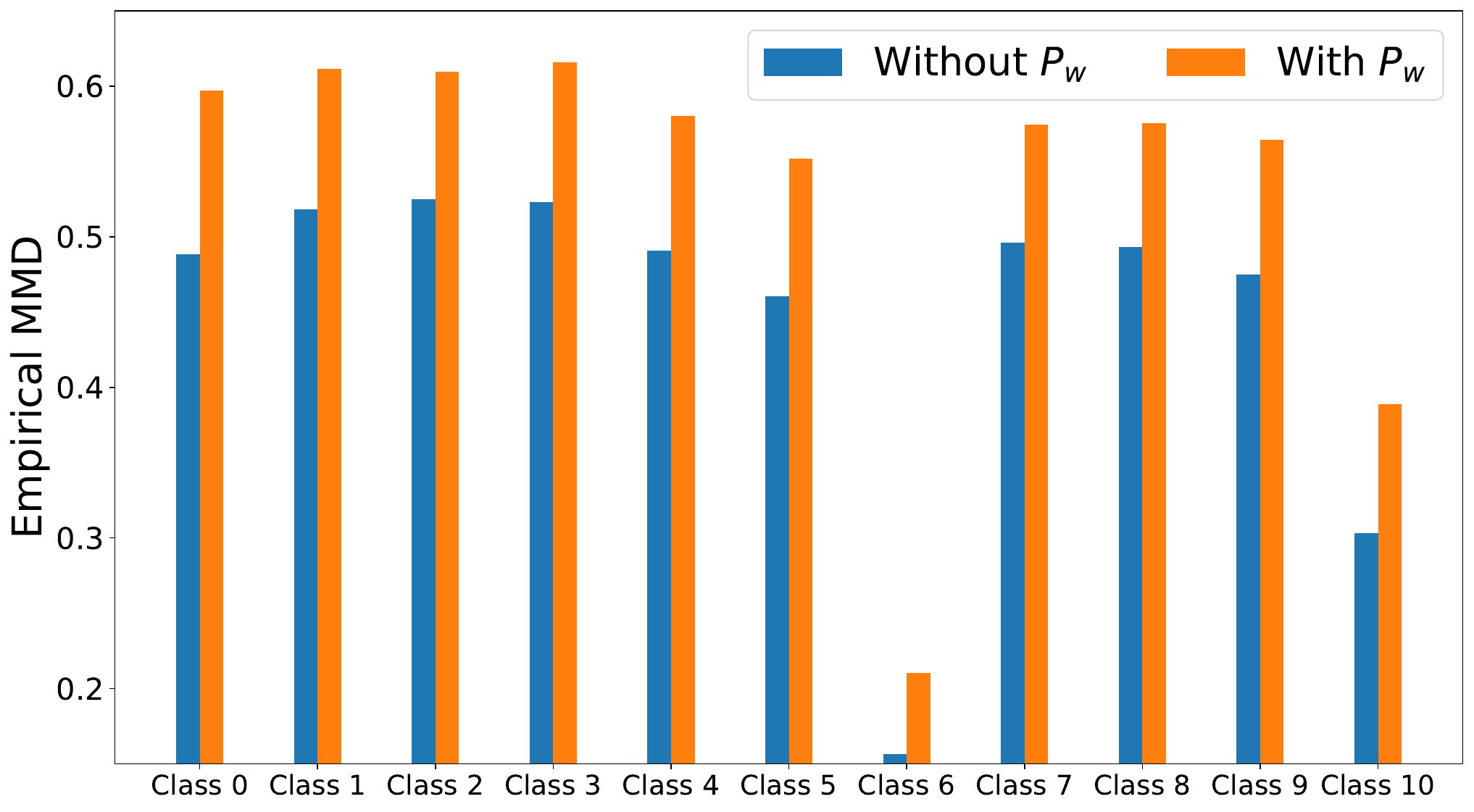} 
\caption{We measure the average MMD of condensed knowledge class-wisely between adjacent rounds on the OrganC dataset. Greater MMD indicates a larger difference in distribution. The vanilla selection would cause more knowledge repetition between rounds. With our model-guided selection ($P_w$) in each round, the condensed knowledge between adjacent rounds is more different.}
\label{fig:organc_mmd_gap}
\end{figure}

\paragraph{Problems.} Most data-centric federated learning methods face problems in the quality and value of the synthesized data, especially under non-IID scenarios. We summarize the problems as follows:
\begin{itemize}
\item Low synthesized data quality. The quality of synthesized and shared virtual data would be low, leading to the model being biased or degraded. For example, the generative models~\cite{goodfellow2020generative} generated data only contain information of the single sample, and the training of generative models with limited local data would be costly and unstable~\cite{li2024feature}. The mix-up of the real data would distort the data due to the non-linearity between input space and label space, causing incorrect data pairs~\cite{verma2019manifold}. The quality of model inversion is highly related to the model performance. Biased models under non-IID scenarios fail to generate data of fidelity. The distribution matching would distill knowledge from local data. However, it only match the final features and it still faces problems dealing with extracting subtle and discriminative features from larger image sizes with less inter-class difference.
\item Repeated knowledge. Despite one can synthesize data with high fidelity, its benefits for current models are under-explored. Without explicitly considering the current global model, the knowledge in synthesized data tends to be repeated and homogeneous in each round, which would be less effective after several rounds. The global model thus can hardly advance its performance.
\item Privacy risk. Some synthesis methods would easily unveil the privacy of local data. The synthetic data should carry minimum irrelevant and individual information. However, the real data mixture and data generated by generative models are highly similar to the private data, and broadcast or peer-to-peer communication~\cite{huang2024overcoming} increases the security and privacy concerns.
\end{itemize}
We summarize the problems faced by representative methods in Table~\ref{tab:summary_chanllenge}. Moreover, most data-centric federated learning methods still rely on frequent communication, thus they are not communication efficient.

\begin{figure}[ht]
\centering
\includegraphics[width=0.8\columnwidth]{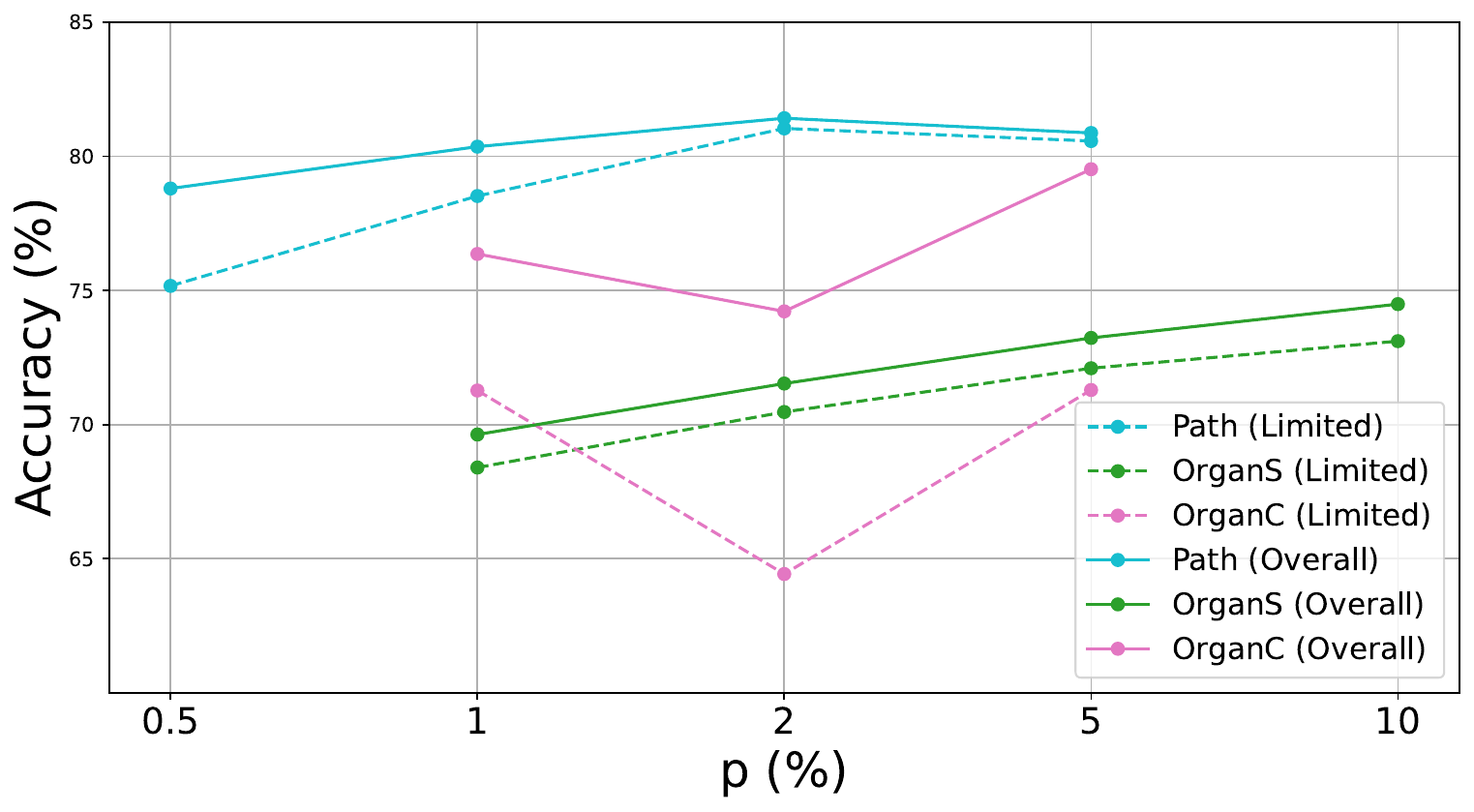} 
\caption{Predictive performance under various $p$ values on medical datasets. `Limited' indicates the performance under limited 10 communication rounds. `Overall' indicates the overall performance with adequate communication rounds.}
\label{fig:p_medical}
\end{figure}

\begin{figure}[ht]
\centering
\includegraphics[width=0.8\columnwidth]{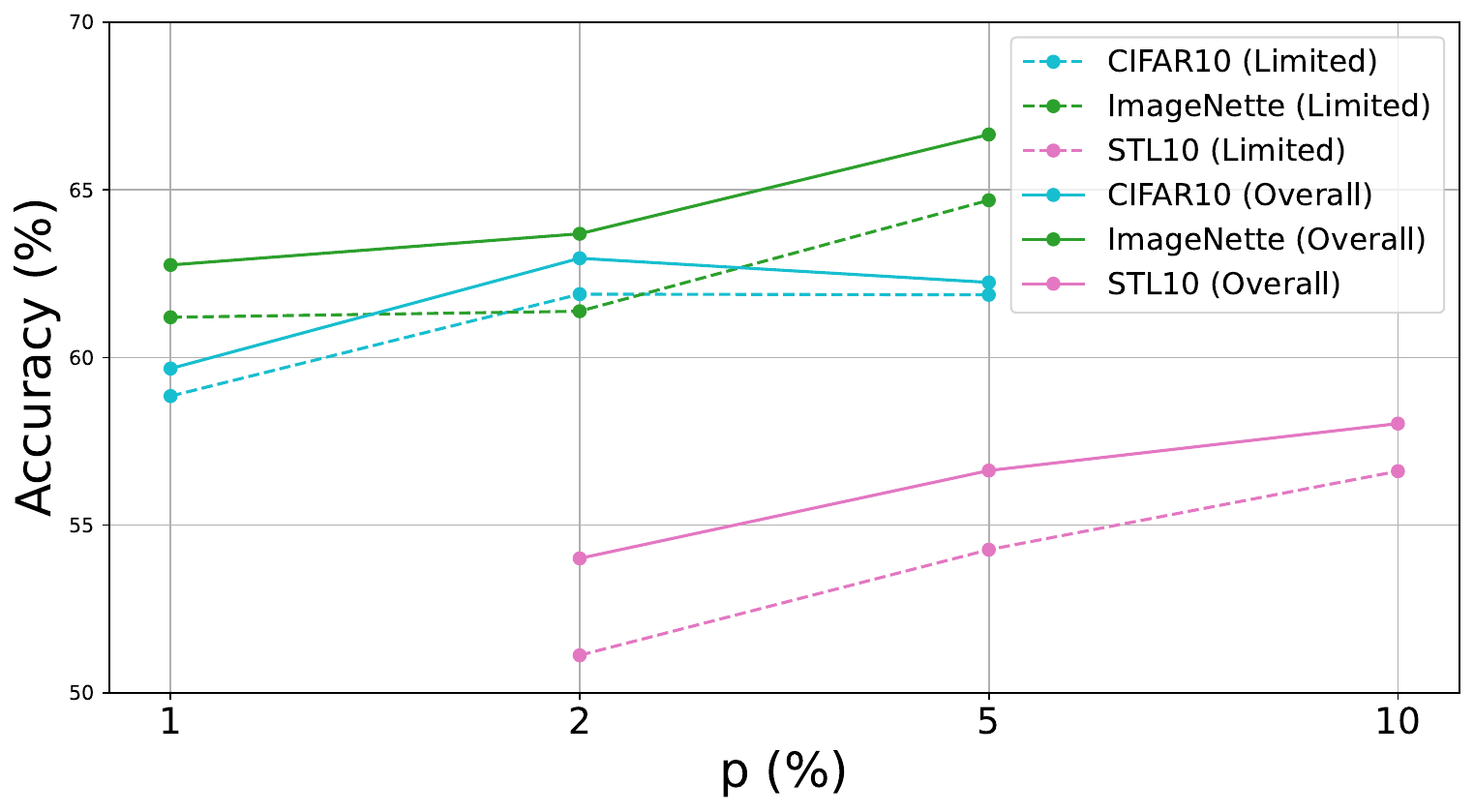} 
\caption{Predictive performance under various $p$ values on natural datasets. `Limited' indicates the performance under limited communication rounds (10). `Overall' indicates the overall performance with adequate communication rounds.}
\label{fig:p_natural}
\end{figure}

\begin{table*}[htbp]
\centering
\small
\setlength{\tabcolsep}{1mm}
\begin{tabular}{cc|ccccc|ccc}
\hline
\multicolumn{2}{c|}{\textbf{Method}}   & \textbf{Path} & \textbf{OCT} & \textbf{OrganS} & \textbf{OrganC} &\textbf{Pneumonia} & \textbf{CIFAR10} & \textbf{STL10} & \textbf{ImageNette} \\ \hline
\multicolumn{2}{c|}{FedMix, DFRD} & \multirow{4}{*}{12.13 MB} & \multirow{4}{*}{11.65 MB} & \multirow{4}{*}{12.20 MB} & \multirow{4}{*}{426.15 MB} & \multirow{4}{*}{426.15 MB} & \multirow{4}{*}{12.21 MB} & \multirow{4}{*}{12.21 MB} & \multirow{4}{*}{14.55 MB} \\ 
\multicolumn{2}{c|}{FedAvg, FedProx} &  & & & & & & & \\ 
\multicolumn{2}{c|}{MOON, FedGen} &  &  & & &  & & & \\ 
\multicolumn{2}{c|}{DESA} & & &  & &  & & & \\ \hline
\multicolumn{2}{c|}{FedGAN} & 178.85 MB & 178.13 MB & 178.69 MB & 2349.40 MB & 2349.40 MB & 178.93 MB & 178.93 MB & 276.20 MB \\ \hline
\multicolumn{2}{c|}{\sys, FedDM} & \textbf{2.04 MB} & \textbf{0.74 MB} & \textbf{0.52 MB} & \textbf{30.72 MB} & \textbf{11.77 MB} & \textbf{2.96 MB} & \textbf{0.25 MB} & \textbf{1.48 MB} \\
\multicolumn{2}{c|}{$p$\%} & 1\% & 1\% & 5\% & 5\% & 5\% & 2\% & 5\% & 1\% \\ \hline
\end{tabular}
\caption{Per-round upload communication costs of all clients. $p$\% indicates the size of condensed knowledge dataset as a proportion of the size of the original dataset.}
\label{tab:full_commnucation_msg_quatity}
\end{table*}

\begin{table}[htbp]
\centering
\setlength{\tabcolsep}{1mm}
\begin{tabular}{c|c|c|c}
\hline
     & \textbf{Initialization} & \textbf{Upload} & \textbf{Download} \\ \hline
FedMix & Mix up data & \multirow{7}{*}{Model} & \multirow{6}{*}{Model}\\ \cline{2-2}
DSEA & Anchor data & & \\ \cline{2-2}
FedAvg & \multirow{8}{*}{None} &  &  \\ 
FedProx & &  &  \\ 
MOON & &  &  \\ 
DRFD &  &  &  \\ \cline{4-4}
FedGen & & & Model, G \\ \cline{3-4}
FedGAN & & Model, G, D & Model, G, D \\ \cline{3-4}
FedDM & & \multirow{2}{*}{Knowledge} & \multirow{2}{*}{Model} \\ 
\sys & & &  \\ \hline
\end{tabular}
\caption{Message types of each method uploaded and downloaded in each round. G denotes the generator and D denotes the discriminator. Initialization indicates the preparation phase before the federated learning.}
\label{tab:commnucation_msg}
\end{table}

\begin{table}[htbp]
\centering
\small
\setlength{\tabcolsep}{1mm}
\begin{tabular}{c|cc|cc|cc}
\hline
Acc(\%) & \multicolumn{2}{c|}{\textbf{CIFAR10}} & \multicolumn{2}{c|}{\textbf{STL10}} & \multicolumn{2}{c}{\textbf{ImageNette}} \\ \hline
$\beta$ & 0.05 & 0.02 & 0.05 & 0.02 & 0.05 & 0.02  \\
\hline
FedAvg & 38.68 & 26.82 & 33.09 & 27.18 & 34.22 & 11.75  \\
FedProx & 37.58 & 33.67 &  28.58 & 27.91 & 42.42 & 22.78 \\
MOON & 36.20 & 28.06 & 29.95 & 28.95 & 22.11 & 10.83 \\ \hline
FedGen & 24.50 & 24.37 & 26.40 & 26.44 & 36.19 & 27.35 \\
FedGAN & 36.63 & 31.20 & 25.90 & 29.28 & 44.18 & 25.86 \\
FedMix & 37.97 & 31.83 & 27.25 & 27.01 & 40.82 & 24.23 \\
DFRD & 35.98 & 27.65 & 25.07 & 19.26 & 27.01 & 14.47 \\
FedDM & 54.62 & 47.96 & 53.70 & 49.71 & 50.39 & 39.97 \\
DESA & 51.75 & 47.03 & 42.83 & 35.86 & 40.79 & 24.87 \\ \hline
\textbf{\sys}  & \textbf{61.89} & \textbf{60.04} & \textbf{54.91} & \textbf{55.04} & \textbf{61.20} & \textbf{57.94}\\
\hline
\end{tabular}
\caption{Predictive accuracy comparison on natural datasets under two non-IID scenarios: $Dir(0.05)$ and $Dir(0.02)$ under limited communication budgets. We adopt the ConvNet model by default. \textbf{Bold} numbers indicate the best accuracy results.}
\label{tab:limited_natural}
\end{table}


\begin{table}[htbp]
\centering
\small
\setlength{\tabcolsep}{1mm}
\begin{tabular}{c|c|c|c}
\hline
\textbf{Method} & \textbf{Max AUC $\downarrow$} & \textbf{Mean AUC $\downarrow$} & \textbf{Defense Rate $\uparrow$}  \\ \hline
\textbf{FedAvg} & 0.667 & 0.575 & 10\%  \\
\textbf{\sys} & \textbf{0.544} & \textbf{0.514} & \textbf{30\%} \\ \hline

\end{tabular}
\caption{The AUC results of MIA on CIFAR10 dataset under $Dir(0.05)$. $\downarrow$ means the lower, the better. $\uparrow$ means the opposite.}
\label{tab:full MIA results}
\end{table}

\begin{table}[ht]
\centering
\small
\setlength{\tabcolsep}{1mm}
\begin{tabular}{c|cccc}
\hline
\textbf{$K$} & 3 & 5 & 7 & 9 \\
\hline
\textbf{Path} (nclass=9) & 76.25 & \textbf{78.52} & 77.58 & 75.77 \\
\hline
\textbf{OrganS} (nclass=11) & 72.24 & \textbf{72.65} & 72.09 & 71.92 \\
\hline
\end{tabular}
\caption{Ablation Study of $K$ value}
\label{tab:abla_k}
\end{table}

\section{Impact of $p$ Value}
In experiments we set the size of the condensed knowledge dataset to $p$\% of the size of the original local datasets, where the value of $p$ is selected from \{1,2,5\}. Intuitively, larger $p$ indicates that knowledge dataset can have larger capacity to contain more knowledge. But it also increase the difficulty of optimization and increasing costs in computation and communication. To evaluate the impact of the selection of $p$, we test the both overall predictive performance and performance under limited communication rounds under wider range of $p$ values selected from 0.5 to 10. The experimental results on medical datasets and natural datasets are shown in Figure~\ref{fig:p_medical} and Figure~\ref{fig:p_natural}. Generally, with larger size of knowledge size, the predictive performance would be better, but there is still a law of diminishing marginal utility with increasing overhead of optimization and communication.

\section{Ablation Study of $K$ Value in Eq.~13}
We conduct the ablation study of the selection of $K$ in Eq.~13. We select different $K$ and test the performance on Path and OrganS datasets under limited communication budgets. The experimental results are shown in Table~\ref{tab:abla_k}. We conclude that $K$ should be chosen as a moderate value relative to the total number of classes. A small $K$ would result in an insufficient number of negative samples, while a large $K$ may introduce unnecessary and noisy `easy negative' samples.

\begin{table}[ht]
\centering
\small
\setlength{\tabcolsep}{1mm}
\begin{tabular}{c|cc|cc}
\hline
Acc(\%) & \textbf{Path} & \textbf{OrganS} & \textbf{CIFAR10} & \textbf{ImageNette}  \\ 
\hline
w.o. all & 72.76 & 71.32 & 54.62 & 50.39  \\
w.o. $L_{rc}$ + $P_w$ & 73.04 & 71.74 & 60.54 &  57.99    \\ 
w.o. $L_{rc}$ & 74.52 & 71.93  & 61.19 &  60.42      \\ \hline
\textbf{\sys} & 78.52 & 72.65 & 61.89 & 61.20 \\
\hline
w.o. all & 73.97 & 71.37 & 54.75 & 52.25  \\
w.o. $L_{rc}$ + $P_w$ & 76.48 & 72.13 & 61.35 & 60.92  \\ 
w.o. $L_{rc}$ & 78.56 & 72.40  & 61.69 &  61.94      \\ \hline
\textbf{\sys} & 80.36 & 73.23 & 62.95 & 62.76 \\
\hline
\end{tabular}
\caption{Ablation study on both medical and natural datasets. The top sub-table is the performance under limited communication budgets and the bottom sub-table is the performance under adequate communication budgets.}
\label{tab:full_ablation_study}
\end{table}

\section{More Communication Analysis}
We analyze and categorize the message types of baseline methods and our method during the federated learning process in Table~\ref{tab:commnucation_msg}. Initialization indicates the preparation phase before federated learning. Our method does not need to communicate before and enjoys a small dataset-level rather than model-level upload costs. The ConvNet model and ResNet18 model in our experiments have around 3.2$\times$10$^5$ and 1.12$\times$10$^7$ parameters with float32 precision, respectively. Thus, each parameter would take 4 bytes. For our condensed knowledge dataset, we can adopt the PIL format where one pixel takes 1 byte to store and transmit.

\section{More Experiments about Model-guided Selection}
Besides the illustration on Path dataset in the main content, we also conduct analytic experiments on the OrganC dataset. we measure the empirical MMD of the condensed knowledge between two adjacent rounds class-wisely with or without model-guided selection in Figure~\ref{fig:organc_mmd_gap}. Empirically, the larger the MMD indicates, the more different the distributions are. We can note that with model-guided selection (with $P_w$), the condensed knowledge between two adjacent rounds shows a higher MMD value. It reflects that it can force to condense more different knowledge from the previous round and thus avoid unnecessary repetition. The heterogeneous and model-specific knowledge would be more effective and valuable for model updating.

\section{Experiments on Natural Datasets}
We compare our methods with baseline methods on natural datasets. The experimental results of the overall predictive performance are shown in Table~\ref{tab:adequate_natural}. The experimental results of the predictive performance under limited communication budgets are shown in Table~\ref{tab:limited_natural}. In addition to predictive performance, we conduct an ablation study in Table~\ref{tab:full_ablation_study} and analyze upload communication costs in Table~\ref{tab:full_commnucation_msg_quatity} on natural datasets. We also conduct the MIA experiment to test the privacy-preserving ability on natural datasets in Table~\ref{tab:full MIA results}. All of these experiments demonstrate the advantages of our methods in terms of predictive performance, communication costs, and privacy preservation ability.

\section{Training Cost Analysis}
In each client, we need to optimize $O(\frac{pNwd}{100M})$ parameters and perform forward and backward $O(T)$ times for condensation, where $N$ is the size of the dataset, $\frac{p}{100}$ is the total size of condensed knowledge, $M$ is the number of clients. $p\%$ is a small value.

\section{More Privacy Analysis}
We initialize the knowledge dataset $S$ with $\mathcal{N}(0,1)$. Knowledge can be condensed by matching the latent feature distributions with constraints. \cite{dong2022privacy} has demonstrated that each sample in the selected batch contributes equally in condensation and on one is dominated. Thus, it is difficult to infer individual privacy. \cite{dong2022privacy} also connects to differential privacy~\cite{dwork2006calibrating} and claims that the privacy budget $\epsilon$ is of the order of $\mathcal{O}(\frac{|\mathcal{S}|}{|\mathcal{T}|})$, which is a small value because $|\mathcal{S}| \ll |\mathcal{T}|$.



\end{document}